\documentclass[letterpaper, 10 pt, conference]{ieeeconf}  

\IEEEoverridecommandlockouts                              

\usepackage{amsmath} 
\usepackage{amssymb}  

\usepackage{graphicx}
\usepackage[font=normalsize]{caption}
\usepackage{hyperref}
\usepackage{cleveref}
\usepackage{booktabs}
\usepackage{multirow}
\usepackage{footnote}
\usepackage{combelow}

\title{\LARGE \bf
Goal-Conditioned End-to-End Visuomotor Control \\ for Versatile Skill Primitives
}

\author{Oliver Groth$^{1}$, Chia-Man Hung$^{1}$, Andrea Vedaldi$^{1}$, Ingmar Posner$^{1}$
\thanks{*This work is supported by ERC 638009-IDIU and EPSRC EP/M019918/1.}
\thanks{$^{1}$Department of Engineering Science,
        University of Oxford, United Kingdom.
        Corresponding author: {\tt\small ogroth@robots.ox.ac.uk}}%
}

\begin{document}

\maketitle
\thispagestyle{empty}
\pagestyle{empty}

\begin{abstract}
Visuomotor control (VMC) is an effective means of achieving basic manipulation tasks such as pushing or pick-and-place from raw images.
Conditioning VMC on desired goal states is a promising way of achieving versatile \emph{skill primitives}.
However, common conditioning schemes either rely on task-specific fine tuning - e.g.~using one-shot imitation learning (IL) - or on sampling approaches using a forward model of scene dynamics i.e.~model-predictive control (MPC), leaving deployability and planning horizon severely limited.
In this paper we propose a conditioning scheme which avoids these pitfalls by learning the controller and its conditioning in an end-to-end manner.
Our model predicts complex action sequences based directly on a dynamic image representation of the robot motion and the distance to a given target observation.
In contrast to related works, this enables our approach to efficiently perform complex manipulation tasks from raw image observations without predefined control primitives or test time demonstrations.
We report significant improvements in task success over representative MPC and IL baselines.
We also demonstrate our model's generalisation capabilities in challenging, unseen tasks featuring visual noise, cluttered scenes and unseen object geometries.
\end{abstract}

\section{Introduction}\label{sec:intro}

\begin{figure*}[ht!]
    \centering
    \includegraphics[width=.95\textwidth]{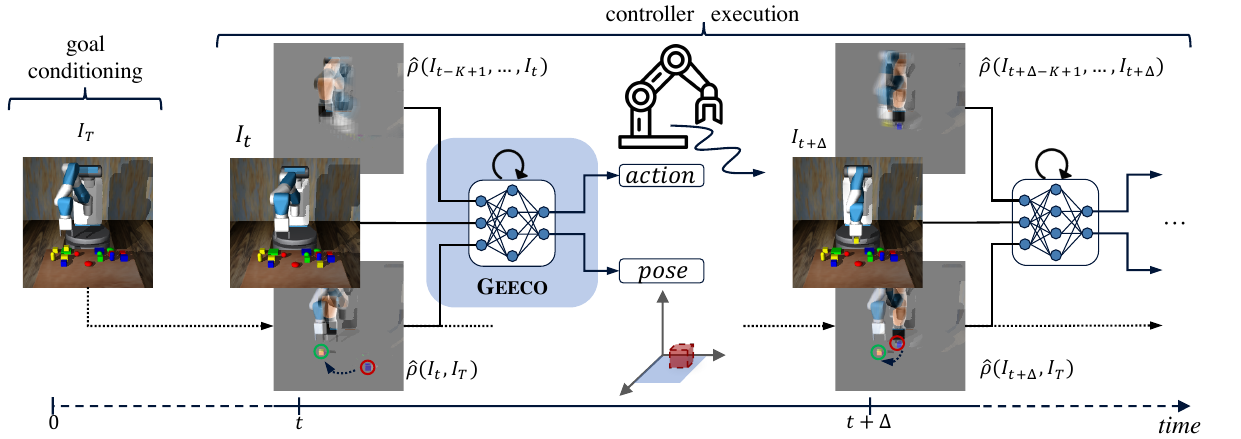}
    \caption{
    Our proposed model executes a task given by a target image $I_T$.
    In this example, $I_T$ indicates that the small yellow cube from the front right needs to be moved onto the green pad in the back left.
    Dynamic images are used to (1) represent the difference between the current observation and the target $\hat{\rho}(I_t,I_T)$ and (2) to capture the motion dynamics $\hat{\rho}(I_{t-K+1},\dots,I_t)$.
    Current and target location of the manipulated object are highlighted by red and green circles respectively.
    }
    \label{fig:teaser}
\end{figure*}

With recent advances in deep learning, we can now learn robotic controllers end-to-end, mapping directly from raw video streams into a robot's command space.
The promise of these approaches is to build real-time visuomotor controllers without the need for complex pipelines or predefined macro-actions (e.g.~for grasping).
End-to-end visuomotor controllers have demonstrated remarkable performance in real systems, e.g.~learning to pick up a cube and place it in a basket~\cite{james2017corl,zhu2018rss}.
However, a common drawback of current visuomotor controllers is their limited \emph{versatility} due to an often very narrow task definition.
For example, in the controllers of~\cite{james2017corl,zhu2018rss}, which are unconditioned, putting a red cube into a blue basket is a different task than putting a yellow cube into a green basket.
In contrast to that, in this paper we consider a broader definition of task and argue that it should rather be treated as a \emph{skill primitive} (e.g.~a policy which can pick up any object and place it anywhere else).
Such a policy must thus be conditioned on certain arguments, e.g.~specifying the object to be moved and its target.

Several schemes have been proposed to condition visuomotor controllers on a \emph{target image}, e.g.~an image depicting how a scene should look like after the robot has executed its task.
Established conditioning schemes build on various approaches such as model-predictive control~\cite{ebert2018arxiv}, task-embedding~\cite{james2018corl} or meta-learning~\cite{finn2017corl} and are discussed in greater detail in \cref{sec:related}.
However, the different methods rely on costly sampling techniques, access to prior demonstrations or task-specific fine-tuning during test time restraining their general applicability.

In contrast to prior work, we propose an efficient end-to-end controller which can be conditioned on a single target image without fine-tuning and regresses directly to motor commands of an actuator without any predefined macro-actions.
This allows us to learn general \emph{skill primitives}, e.g.~pushing and pick-and-place skills, which are versatile enough to immediately generalise to new tasks, i.e.~unseen scene setups and objects to handle.
Our model utilises \emph{dynamic images}~\cite{bilen2017ieee} as a succinct representation of the video dynamics in its observation buffer as well as a visual estimation of the difference between its current observation and the target it is supposed to accomplish.
\Cref{fig:teaser} depicts an example execution of our visuomotor controller, its conditioning scheme and intermediate observation representations.

In summary, our contributions are three-fold: Firstly, we propose a novel architecture for visuomotor control which can be efficiently conditioned on a new task with just one single target image.
Secondly, we demonstrate our model's efficacy by outperforming representative MPC and IL baselines in pushing and pick-and-place tasks by significant margins.
Lastly, we analyse the impact of the dynamic image representation in visuomotor control providing beneficial perception invariances to facilitate controller resilience and generalisation without the need of sophisticated domain randomisation schemes during training.

\section{Related Work}\label{sec:related}
The problem of \emph{goal conditioning} constitutes a key challenge in visuomotor control: Given a specific task specification (e.g.~putting a red cube onto a blue pad), it needs to be communicated to the robot, which in turn must adapt its control policy in such a way that it can carry out the task.
In this paper, we focus on goals which are communicated visually, i.e.~through images depicting how objects should be placed on a table.
Prior methods which have shown impressive real-world results typically involve dedicated sub-modules for perception and planning~\cite{labbe2020mcts} or are only loosely goal-conditioned, e.g. on a target shape category~\cite{pashevich2020shapecategories}.
We restrict our survey to end-to-end controllers which can be conditioned on a \emph{single target image} and group related work by their condition schemes and action optimisation methods.

In \emph{visual model-predictive control} one learns a forward model of the world, forecasting the outcome of an action.
The learned dynamics model is then explored via sampling or gradient-based methods to compute a sequence of actions which brings the predicted observation closest to a desired goal observation.
An established line of work on \emph{Deep Visual Foresight} (VFS)~\cite{finn2017icra,ebert2018arxiv,nair2019arxiv,xie2019arxiv} learns action-conditioned video predictors and employs CEM-like~\cite{rubinstein2004springer} sampling methods for trajectory optimisation, successfully applying those models to simulated and real robotic pushing tasks.
Instead of low-level video prediction, visual MPC can also be instantiated using higher-level, object-centric models for tasks such as block stacking~\cite{ye2020objectfw,janner2019reasoning}.
Another line of work attempts to learn forward dynamics models in suitable latent spaces.
After projecting an observation and a goal image into the latent space, a feasible action sequence can then be computed using gradient-based optimisation methods~\cite{watter2015nips,byravan2018icra,srinivas2018icml,yu2019rss}.
Even though MPC approaches have shown promising results in robot manipulation tasks, they are limited by the quality of the forward model and do not scale well due to the action sampling or gradient optimisation procedures required.
In contrast to them our model regresses directly to the next command given a buffer of previous observations.

\emph{One-Shot Imitation Learning} seeks to learn general task representations which are quickly adaptable to unseen setups.
MIL~\cite{finn2017corl} is a meta-controller, which requires fine-tuning during test time on one example demonstration of the new task to adapt to it.
In contrast to MIL, TecNet~\cite{james2018corl} learns a task embedding from expert demonstrations and requires at least one demonstration of the new task during test time to modulate its policy according to similar task embeddings seen during training.
Additionally, a parallel line of work in that domain operates on discrete action spaces~\cite{xu2018icra,huang2019cvpr} and maps demonstrations of new tasks to known macro actions.
Unlike those methods, our model is conditioned on a single target image and does not require any fine-tuning on a new task during test time.

\emph{Goal-conditioned reinforcement learning}~\cite{kaelbling1993goalrl} is another established paradigm for learning of control policies.
However, due to the unwieldy nature of images as state observations, the use of goal images typically comes with limiting assumptions such as being from a previously observed set~\cite{warde-farley2018iclr} or living on the manifold of a learned latent space~\cite{nair2018visual}.
Our proposed utilisation of dynamic images for goal conditioning circumvents such limitations and can be seen as complementary to other work which incorporates demonstrations into goal-conditioned policy learning~\cite{nair2018icra,ding2019neurips} by enabling efficient bootstrapping of a control policy on goal-conditioned demonstrations.

\section{Goal-Conditioned Visuomotor Control}\label{sec:method}
In order to build a visuomotor controller which can be efficiently conditioned on a target image and is versatile enough to generalise its learned policy to new tasks immediately, we need to address the following problems:
Firstly, we need an efficient way to detect scene changes, i.e.~answering the question `Which object has been moved and where from and to?'
Secondly, we want to filter the raw visual observation stream such that we only retain information pertinent to the control task; specifically the motion dynamics of the robot.
Drawing inspiration from previous work in VMC and action recognition, we propose \emph{GEECO}, a novel architecture for \emph{goal-conditioned end-to-end control} which combines the idea of \emph{dynamic images}~\cite{bilen2017ieee} with a robust end-to-end controller network~\cite{james2017corl} to learn versatile manipulation skill primitives which can be conditioned on new tasks on the fly.
We discuss next the individual components.

\begin{figure}[t!]
    \centering
    \includegraphics[width=.9\columnwidth]{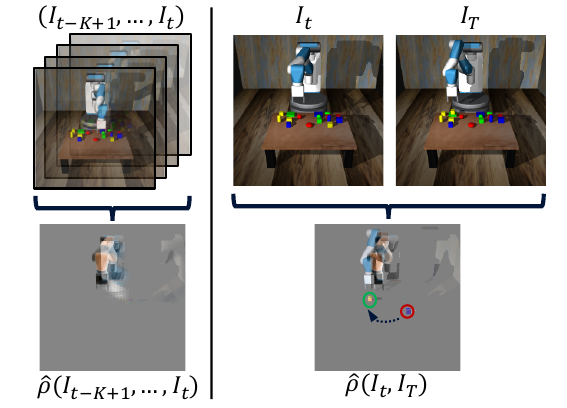}
    \caption{
    Utilisation of dynamic images.
    Left: A dynamic image represents the motion occurring in a sequence of $K$ consecutive RGB observations.
    Right: A dynamic image represents the changes which occurred between the two images $I_t$ and $I_T$ like the change of object positions as indicated by the red and green circles.
    }
    \label{fig:method-dynimg}
\end{figure}

\textbf{Dynamic Images.}
In the domain of action recognition, dynamic images have been developed as a succinct video representation capturing the dynamics of an entire frame sequence in a single image.
This enables the treatment of a video with convolutional neural networks as if it was an ordinary RGB image facilitating dynamics-related feature extraction.
The core of the dynamic image representation is a ranking machine which learns to sort the frames of a video temporally~\cite{fernando2015cvpr}.
As shown by prior work~\cite{bilen2017ieee}, an approximate linear ranking operator $\hat{\rho}(\cdot)$ can be applied to any sequence of $H$ temporally ordered frames $(I_1, \dots, I_H)$ and any image feature extraction function $\psi(\cdot)$ to obtain a dynamic feature map according to the following \cref{eq:dynimg-rhohat}:
\begin{align}
\hat{\rho}(I_1,\dots,I_H;\psi) = \sum_{t=1}^H{\alpha_t\psi(I_t)}
\label{eq:dynimg-rhohat}
\\
\alpha_t = 2(H-t+1) - (H+1)(\mathcal{H}_H - \mathcal{H}_{t-1})
\label{eq:dynimg-alpha}
\end{align}
Here, $\mathcal{H}_t=\sum_{i=1}^t{1/t}$ is the $t$-th Harmonic number and $\mathcal{H}_0=0$.
Setting $\psi(\cdot)$ to the identity, $\hat{\rho}(I_1, \dots, I_H)$ yields a dynamic image which, after normalisation across all channels, can be treated as a normal RGB image by a downstream network.
The employment of dynamic images serves two important purposes in our network, as depicted in \cref{fig:method-dynimg}.
Firstly, it compresses a window of the last $K$ RGB observations into one image $\hat{\rho}(I_{t-K+1},\dots,I_t)$ capturing the current motion of the robot arm.
Secondly, given a target image $I_T$ depicting the final state the scene should be in, the dynamic image $\hat{\rho}(I_t, I_T)$ lends itself very naturally to represent the \emph{visual difference} between the current observation $I_t$ and the target state $I_T$.
Another advantage of using dynamic images in these two places is to make the controller network invariant w.r.t.~the static scene background and, approximately, the object colour, allowing it to focus on location and geometry of objects involved in the manipulation task.

\textbf{Observation Buffer.}
During execution, our network maintains a buffer of most recent $K$ observations as a sequence of pairs $\left ( (I_{t-K+1}, \mathbf{x}_{t-K+1}), \dots, (I_t, \mathbf{x_t}) \right )$ where $I_t$ is the RGB frame at time step $t$ and $\mathbf{x_t}$ is the proprioceptive feature of the robot at the same time step represented as a vector of its joint angles.
Throughout our experiments we set $K=4$.
The observation buffer breaks long-horizon manipulation trajectories into shorter windows which retain relative independence from each other.
This endows the controller with a certain error-correction capacity, e.g.~when a grasped object slips from the gripper prematurely, the network can regress back to a pick-up phase.

\textbf{Goal Conditioning.}
Before executing a trajectory, our controller is \emph{conditioned} on the task to execute, i.e.~moving an object from its initial position to a goal position, via a target image $I_T$ depicting the scene after the task has been carried out.
As shown in \cref{fig:method-dynimg} (right), the dynamic image representation $\hat{\rho}(I_t,I_T)$ helps this inference process by only retaining the two object positions and the difference in the robot pose while cancelling out all static parts of the scene.

\begin{figure}[t!]
    \centering
    \includegraphics[width=\columnwidth]{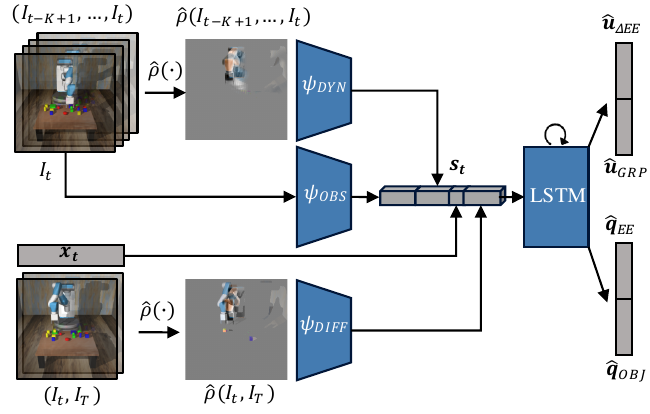}
    \caption{
    Network architecture of GEECO-$\mathcal{F}$.
    The observation buffer $(I_{t-K+1}, \dots, I_T)$ is compressed into a dynamic image via $\hat{\rho}(.)$ and passed through $\psi_{DYN}$.
    The current difference to the target frame $I_T$ is also computed via $\hat{\rho}(.)$ and passed through $\psi_{DIFF}$.
    Lastly, the current observation $I_t$ is encoded via $\psi_{OBS}$.
    All CNNs compute spatial feature maps which are concatenated to the tiled proprioceptive feature $\mathbf{x}_t$.
    The LSTM's output is decoded into command actions $\mathbf{\hat{u}}_{\Delta EE}$ and $\mathbf{\hat{u}}_{GRP}$ as well as auxiliary pose predictions $\mathbf{\hat{q}}_{EE}$ and $\mathbf{\hat{q}}_{OBJ}$.
    }\label{fig:method-model}
\end{figure}

\textbf{Network Architecture.}
The controller network takes the current observation buffer $\left ( (I_{t-K+1}, \mathbf{x}_{t-K+1}), \dots, (I_t, \mathbf{x}_t) \right )$ and the target image $I_T$ as input and regresses to the following two action outputs:
(1) The change in Cartesian coordinates of the end effector $\mathbf{\hat{u}}_{\Delta EE}$ and
(2) a discrete signal $\mathbf{\hat{u}}_{GRP}$ for the gripper to either open $(-1)$, close $(+1)$ or stay in position $(0)$.
Additionally, the controller regresses two auxiliary outputs:
the current position of the end effector $\mathbf{\hat{q}}_{EE}$ and of the object to manipulate $\mathbf{\hat{q}}_{OBJ}$, both in absolute Cartesian world coordinates.
While the action vectors $\mathbf{\hat{u}}_{\Delta EE}$ and $\mathbf{\hat{u}}_{GRP}$ are directly used to control the robot, the position predictions serve as an auxiliary signal during the supervised training process to encourage the network to learn intermediate representations correlated to the world coordinates.
A sketch of our model architecture can be found in \cref{fig:method-model} and a detailed description of all architectural parameters can be found in~\cref{sec:geeco-hparams}.

The full model is trained in an end-to-end fashion on $N$ expert demonstrations of manipulation tasks collected in a simulation environment.
Each expert demonstration is a sequence of $H$ time steps indexed by $t$ containing:
the RGB frame $I_t$, the proprioceptive feature $\mathbf{x}_t$, the robot commands $\mathbf{u}_{\Delta EE}^*(t)$, $\mathbf{u}_{GRP}^*(t)$ and the positions of the end effector and the object to manipulate $\mathbf{q}_{EE}^*(t)$, $\mathbf{q}_{GRP}^*(t)$.
During training, we minimise the following loss function:
\begin{equation}
\begin{split}
    \mathcal{L} = \sum_{i=1}^N
    \Bigg[
    \sum_{t=1}^{H-K+1}
     & \operatorname{MSE}(\mathbf{\hat{u}}_{\Delta EE}(\mathbf{\tau}_{i,t}), \mathbf{u}_{\Delta EE}^*(i,t)) \\
    + & \operatorname{CCE}(\mathbf{\hat{u}}_{GRP}(\mathbf{\tau}_{i,t}), \mathbf{u}_{GRP}^*(i,t)) \\
    + & \lambda \Big( \operatorname{MSE}(\mathbf{\hat{q}}_{EE}(\mathbf{\tau}_{i,t}), \mathbf{q}_{EE}^*(i,t)) \\
    + & \operatorname{MSE}(\mathbf{\hat{q}}_{OBJ}(\mathbf{\tau}_{i,t}), \mathbf{q}_{OBJ}^*(i,t)) \Big)
    \Bigg]
\end{split}
\label{eq:loss-geeco}
\end{equation}
In~\cref{eq:loss-geeco} $\operatorname{MSE}$ and $\operatorname{CCE}$ are abbreviations of \emph{Mean-Squared Error} and \emph{Categorical Cross-Entropy} respectively.
The hyper-parameter $\lambda$ weighs the auxiliary loss terms for pose prediction.
The shorthand notation $\mathbf{\tau}_{i,t}$ represents the $t$-th training window in the $i$-th expert demonstration of the training dataset comprising of $((I_t, \mathbf{x}_t), \dots, (I_{t+K-1}, \mathbf{x}_{t+K-1});~I_T=I_H)$; $\mathbf{u}^*(i,t)$ and $\mathbf{q}^*(i,t)$ are the corresponding ground truth commands, and $\mathbf{\hat{u}}(\mathbf{\tau}_{i,t})$ and $\mathbf{\hat{q}}(\mathbf{\tau}_{i,t})$ are shorthand notations for the network predictions on that window.
During training we always set the target frame $I_T$ to be the last frame of the expert demonstration $I_H$.

\textbf{Model Ablations.}
We refer to our \emph{full} model as GEECO-$\mathcal{F}$ (or just $\mathcal{F}$ for short) as depicted in~\cref{fig:method-model}.
However, in order to gauge the effectiveness of our different architecture design decisions, we also consider two ablations of our full model which are briefly described below.
We refer the reader to~\cref{sec:geeco-ablations} for more comprehensive details and architecture sketches.

GEECO-$\mathcal{R}$:
%
This ablation has $\psi_{DYN}$ and $\psi_{DIFF}$ removed and $\psi_{OBS}$ is responsible for encoding the current observation $I_t$ and the target image $I_T$ and the feature distance is used for goal conditioning.
Thus, the state tensor becomes $\mathbf{s}_t = \psi_{OBS}(I_t) \oplus \mathbf{x}_t \oplus (\psi_{OBS}(I_T) - \psi_{OBS}(I_t))$, where $\oplus$ denotes concatenation along the channel dimension.
This \emph{residual} state encoding serves as a baseline for learning meaningful goal distances in the feature space induced by $\psi_{OBS}$.

GEECO-$\mathcal{D}$:
%
This ablation has only the `motion branch' $\psi_{DYN}$ removed.
The state tensor is comprised of $\mathbf{s}_t = \psi_{OBS}(I_t) \oplus \mathbf{x}_t \oplus \psi_{DIFF}(\hat{\rho}(I_t, I_T))$.
This gauges the effectiveness of using an explicitly shaped goal difference function over an implicitly learned one like in GEECO-$\mathcal{R}$.

\section{Experiments}\label{sec:experiments}
Our experimental design is guided by the following questions:
(1) How do the learned skill primitives compare to representative MPC and IL approaches?
(2) Can our controller deliver on its aspired \emph{versatility} by transferring its skills, acquired only on simple cubes, to novel shapes and adverse conditions?

\begin{figure*}[t]
    \centering
    \includegraphics[width=.92\textwidth]{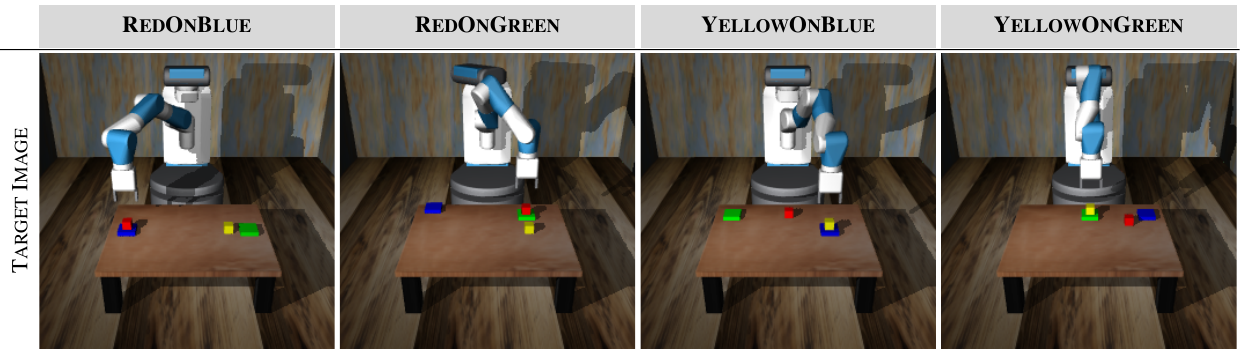}
    \caption{
    Basic manipulation tasks in \textsc{Goal2Cube2}.
    In each task, one of the small cubes (red or yellow) needs to be moved onto one of the target pads (blue or green).
    The tasks can be accomplished via a pushing or pick-and-place manipulation (target pads are reduced to flat textures for pushing).
    }
    \label{fig:exp-scenarios}
\end{figure*}

\textbf{Experimental Setup and Data Collection.}
We have designed a simulation environment, \textsc{Goal2Cube2}, containing four different tasks to train and evaluate our controller on which are presented in~\cref{fig:exp-scenarios}.
In each task one of the small cubes needs to be moved onto one of the larger target pads.
The scenario is designed such that the task is ambiguous and the controller needs to infer the object to manipulate and the target location from a given target image depicting the task to perform.
We use the MuJoCo physics engine~\cite{todorov2012iros} for simulation.
We adapt the Gym environment~\cite{openai2016gym} provided by~\cite{duan2017nips} featuring a model of a \emph{Fetch Mobile Manipulator}~\cite{wise2018fetch} with a 7-DoF arm and a 2-point gripper\footnote{Despite the robot having a mobile platform, its base is fixed during all experiments.}.
For each skill, i.e.~pushing and pick-and-place, we collect 4,000 unique expert demonstrations of the robot completing the task successfully in simulation according to a pre-computed plan.
Each demonstration is four seconds long and is recorded as a list of observation-action tuples at 25 Hz resulting in an episode length of $H = 100$.
%

\begin{figure*}[h]
  \centering
  \includegraphics[width=.92\textwidth]{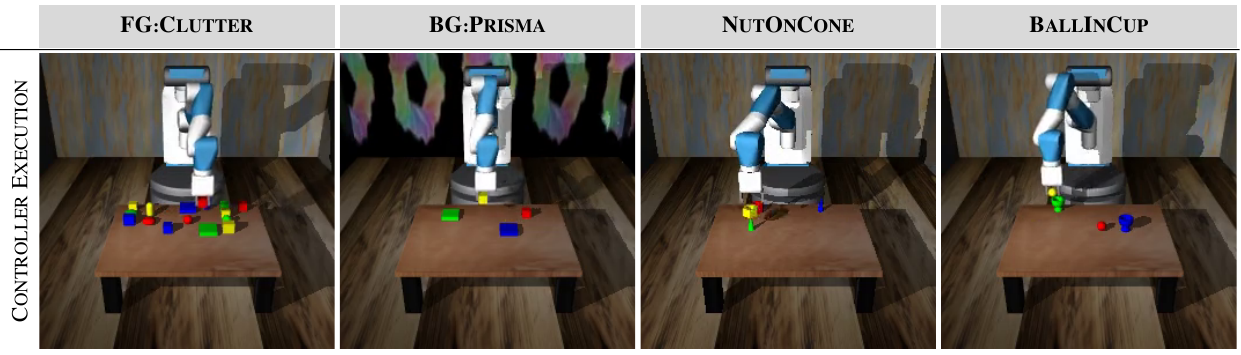}
  \caption{
    Generalisation experiments.
    \textsc{FG:Clutter} and \textsc{BG:Prisma} are variations of \textsc{Goal2Cube2} with severe physical and visual distractions.
    The background in \textsc{BG:Prisma} is a video looping rainbow colour patterns.
    \textsc{NutOnCone} and \textsc{BallInCup} require the application of the pick-and-place skill to novel shapes and target configurations.
  }
  \label{fig:exp-generalisation}
\end{figure*}

\textbf{Baselines for Goal-Conditioned VMC.}
Our first baseline, VFS~\cite{ebert2018arxiv}, is a visual MPC which runs on a video prediction backbone.
A CEM-based~\cite{rubinstein2004springer} action sampler proposes command sequences over a short time horizon which are evaluated by the video predictor.
The sequence which results in a predicted observation closest to the goal image is executed and the process is repeated until termination.
We use SAVP~\cite{lee2018savp} as action-conditioned video predictor and train it on our datasets with the hyper-parameters reported for the BAIR robot-pushing dataset~\cite{ebert2017corl}.
Since the time horizon of our scenarios is much longer than SAVP's prediction horizon, we estimate an upper-bound of the performance of VFS by providing intermediate goal images.\footnote{HVF~\cite{nair2019arxiv} employs a similar `ground truth bottleneck' scheme to upper-bound a visual MPC baseline.}
Our second baseline, \textsc{TecNet}~\cite{james2018corl}, is an IL model which is capable of quick adaptation given the demonstration of a new task.
The demonstration can be as sparse as start and end image of an executed trajectory making it applicable to our setup.
We employ \textsc{TecNet} in its one-shot imitation configuration.
We refer to~\cref{sec:vfs-hparams,sec:tecnet-hparams} for a comprehensive overview over the hyper-parameters used for all baselines.

\textbf{Training and Evaluation Protocol.}
For each skill, we split the demonstrations into training, validation and test sets with a ratio of 2 : 1 : 1 respectively while keeping the task distributions balanced.
We train all models for 300k gradient update steps using the Adam optimiser~\cite{kingma2015iclr}.
After each epoch, we evaluate the controller performance on the validation set and select the best performing model checkpoints for the final evaluation on the test set.

During task execution, we monitor the following performance metrics:
\begin{sc}reach\end{sc} (robot touches object to manipulate at least once)
\begin{sc}pick\end{sc} (robots palm touches object to manipulate while fingers are closed)
\begin{sc}push / place\end{sc} (correct object sits on the target pad by the end of the episode).
Each evaluation episode is terminated after 200 timesteps (8 seconds).

\textbf{Basic Manipulation Results.}
We start our investigation by comparing the performance of our proposed model and its ablations to VFS and \textsc{TecNet} in the \textsc{Goal2Cube2} scenario and report the results in~\cref{table:exp-main}.
We treat the tasks (\textsc{reach}, \textsc{push}, \textsc{pick}, \textsc{place}) as Bernoulli experiments and report their mean success rates for 1,000 trials as well as their binomial proportion confidence intervals at $0.95$.
For pushing tasks we measure nearly perfect reaching and very strong pushing performance for GEECO-$\{\mathcal{R},\mathcal{D},\mathcal{F}\}$ outperforming the baselines by up to $80\%$ on final task success.
We also observe that $\mathcal{D}$ and $\mathcal{F}$, models which employ the dynamic image representation, perform significantly better than the RGB-only ablation $\mathcal{R}$.
Likewise, the general ranking of models also applies to pick-and-place tasks with $\mathcal{D}$ outperforming the baselines by up to $60\%$ task success.
The almost complete failure of VFS and \textsc{TecNet} for a multi-stage task like pick-and-place is unsurprising given that both models have been originally designed for reaching and pushing tasks.
Qualitatively, we observe that VFS reaches the correct object relatively reliably due to its powerful video predictor but struggles to maintain a firm grasp on an object due to its stochastic action sampling.
\textsc{TecNet} fares better in that regard since it is a feedforward regression network like GEECO and can maintain a stable grasp.
However, it often approaches the wrong object to manipulate due to the fact that its policy is modulated by the embedding of similar tasks.
When confronted with subtle task variations like a colour change in a small object the RGB task embedding becomes less informative and \textsc{TecNet} is prone to inferring the wrong task.
An investigation into $\mathcal{F}$'s inferior \textsc{place} performance compared to $\mathcal{D}$ reveals a failure mode of $\mathcal{F}$: The controller sometimes struggles to drop a cube above the target pad presumably due to ambiguity in its depth perception.
This suggests that the signal provided by the dynamic frame buffer at relatively motion-less pivot points can be ambiguous without additional treatment.
When comparing the versions of GEECO which are trained without auxiliary pose supervision ($\lambda = 0.0$) to their fully supervised counterparts ($\lambda = 1.0$), we observe only mild drops in mean performance of up to $15\%$.
Interestingly, the RGB-only ablation $\mathcal{R}$ is least affected by the pose supervision and even improves performance when trained without auxiliary poses.
We hypothesise that this is due to the fact that, in the relatively static simulation, RGB features are very representative of their spatial location.
Generally, we conclude from the pose supervision ablation that GEECO's performance is not entirely dependent on accurate pose supervision enabling it to be trained on even less extensively annotated demonstration data.

\begin{savenotes}
\begin{table*}[ht]
\centering
\vskip 0.15in
\begin{center}
\begin{small}
\begin{sc}
\begin{tabular}{l|cc|ccc}
\toprule
\multirow{2}{*}{} & \multicolumn{2}{c|}{Pushing} & \multicolumn{3}{c}{Pick-and-Place}  \\
& \multicolumn{1}{c}{reach {[}\%{]}} & \multicolumn{1}{c|}{push {[}\%{]}} & \multicolumn{1}{c}{reach {[}\%{]}} & \multicolumn{1}{c}{pick {[}\%{]}} & \multicolumn{1}{c}{place {[}\%{]}} \\
\hline
VFS\footnote{Due to computational constraints, \textsc{VFS} has only been tested on 100 tasks from the test set.}~\cite{ebert2018arxiv} & 66.00 $\pm$ 9.28 & 7.00 $\pm$ 5.00 & 87.00 $\pm$ 6.59 & 40.00 $\pm$ 9.60 & 0.00 $\pm$ 0.00 \\
TecNet~\cite{james2018corl} & 54.10 $\pm$ 3.09 & 15.70 $\pm$ 2.25 & 32.00 $\pm$ 2.89 & 15.70 $\pm$ 2.25 & 0.80 $\pm$ 0.55 \\
\hline
$\mathcal{R}, \lambda=0.0$ & 98.70 $\pm$ 0.70 & 79.80 $\pm$ 2.49 & 86.20 $\pm$ 2.14 & 58.90 $\pm$ 3.05 & 42.50 $\pm$ 3.06 \\
$\mathcal{D}, \lambda=0.0$ & 98.70 $\pm$ 0.70 & \textbf{88.50 $\pm$ 1.98} & \textbf{95.40 $\pm$ 1.30} & 65.80 $\pm$ 2.94 & 54.20 $\pm$ 3.09 \\
$\mathcal{F}, \lambda=0.0$ & 98.00 $\pm$ 0.87 & 78.60 $\pm$ 2.54 & 92.70 $\pm$ 1.61 & 71.00 $\pm$ 2.81 & 45.60 $\pm$ 3.09 \\
\hline
$\mathcal{R}, \lambda=1.0$ & \textbf{98.90 $\pm$ 0.65} & 79.80 $\pm$ 2.49 & 84.90 $\pm$ 2.22 & 50.40 $\pm$ 3.10 & 33.70 $\pm$ 2.93 \\
$\mathcal{D}, \lambda=1.0$ & \textbf{99.30 $\pm$ 0.52} & \textbf{86.60 $\pm$ 2.11} & \textbf{96.20 $\pm$ 1.19} & \textbf{79.90 $\pm$ 2.48} & \textbf{61.40 $\pm$ 3.02} \\
$\mathcal{F}, \lambda=1.0$ & \textbf{99.80 $\pm$ 0.28} & \textbf{89.30 $\pm$ 1.92} & \textbf{94.80 $\pm$ 1.38} & \textbf{78.40 $\pm$ 2.55} & 46.30 $\pm$ 3.09 \\
\bottomrule
\end{tabular}
\end{sc}
\end{small}
\end{center}
\vskip 0.1in
\vspace{-0.1cm}
\caption{
Success rates and confidence intervals for pushing and pick-and-place tasks in the \textsc{Goal2Cube2} scenarios.
}
\label{table:exp-main}
\end{table*}

\begin{table*}[h]
\setlength{\tabcolsep}{4.12pt}
\vskip 0.15in
\begin{center}
\begin{small}
\begin{sc}
\begin{tabular}{l|ccc|ccc}
\toprule
\multirow{2}{*}{} & \multicolumn{3}{c|}{FG:Clutter}  & \multicolumn{3}{c}{BG:Prisma} \\
& reach {[}\%{]} & pick {[}\%{]} & place {[}\%{]} & reach {[}\%{]} & pick {[}\%{]} & place {[}\%{]} \\
\hline
$\mathcal{R}, \lambda=1.0$ & 63.80 $\pm$ 2.98 & 29.40 $\pm$ 2.82 & 15.80 $\pm$ 2.26 & 0.50 $\pm$ 0.44 & 0.10 $\pm$ 0.20 & 0.00 $\pm$ 0.00 \\
$\mathcal{D}, \lambda=1.0$ & 77.00 $\pm$ 2.61 & 42.10 $\pm$ 3.06 & 20.70 $\pm$ 2.51 & \textbf{94.10 $\pm$ 1.46} & \textbf{66.00 $\pm$ 2.94} & \textbf{26.50 $\pm$ 2.74} \\
$\mathcal{F}, \lambda=1.0$ & \textbf{85.50 $\pm$ 2.18} & \textbf{62.60 $\pm$ 3.00} & \textbf{32.60 $\pm$ 2.91} & \textbf{93.40 $\pm$ 1.54} & \textbf{62.40 $\pm$ 3.00} & 19.30 $\pm$ 2.45 \\
\hline
\multirow{2}{*}{} & \multicolumn{3}{c|}{NutOnCone}  & \multicolumn{3}{c}{BallInCup} \\
& reach {[}\%{]} & pick {[}\%{]} & place {[}\%{]} & reach {[}\%{]} & pick {[}\%{]} & place {[}\%{]} \\
\hline
$\mathcal{R}, \lambda=1.0$ & 32.30 $\pm$ 2.90 & 6.90 $\pm$ 1.57 & 0.40 $\pm$ 0.39 & 21.50 $\pm$ 2.55 & 3.90 $\pm$ 1.20 & 0.10 $\pm$ 0.20 \\
$\mathcal{D}, \lambda=1.0$ & \textbf{66.60 $\pm$ 2.92} & \textbf{23.30 $\pm$ 2.62} & 3.30 $\pm$ 1.11 & \textbf{50.60 $\pm$ 3.10} & 9.70 $\pm$ 1.83 & 0.20 $\pm$ 0.28 \\
$\mathcal{F}, \lambda=1.0$ & \textbf{72.20 $\pm$ 2.78} & \textbf{26.90 $\pm$ 2.75} & \textbf{6.20 $\pm$ 1.49} & \textbf{54.60 $\pm$ 3.09} & \textbf{16.30 $\pm$ 2.29} & \textbf{1.50 $\pm$ 0.75} \\
\bottomrule
\end{tabular}

\end{sc}
\end{small}
\end{center}
\vskip 0.1in
\vspace{-0.1cm}
\caption{
Pick-and-place success rates and confidence intervals of GEECO models trained on \textsc{Goal2Cube2} and employed in novel scenarios as depicted in~\cref{fig:exp-generalisation}.
}
\label{table:exp-generalisation}
\end{table*}

\textbf{Generalisation to New Scenarios.}
After validating the efficacy of our proposed approach in basic manipulation scenarios which are close to the training distribution, we investigate its robustness and versatility in two additional sets of experiments.
In the following trials we take models $\mathcal{R}$, $\mathcal{D}$ and $\mathcal{F}$ which have been trained on pick-and-place tasks of \textsc{Goal2Cube2} and apply them to new scenarios probing different aspects of generalisation.
\end{savenotes}  
We present examples of the four new scenarios in~\cref{fig:exp-generalisation} and present quantitative results for 1,000 trials in~\cref{table:exp-generalisation}.
\textsc{FG:Clutter} and \textsc{BG:Prisma} evaluate whether the pick-and-place skill learned by GEECO is robust enough to be executed in visually challenging circumstances as well.
The results reveal that the employment of dynamic images for target difference ($\mathcal{D}$) and additionally buffer representation ($\mathcal{F}$) significantly improves task success over the RGB-baseline ($\mathcal{R}$) in the cluttered tabletop scenario due to the perceptual invariances afforded by the dynamic image representation.
The effect is even more apparent when the colours of the scene background are distorted.
This leads to a complete failure of $\mathcal{R}$ (which is now chasing after flickering colour patterns in the background) while $\mathcal{D}$ and $\mathcal{F}$ can still accomplish the task in about $20\%$ of the cases.
In the second set of experiments (cf.~\cref{table:exp-generalisation}, bottom), we evaluate whether the pick-and-place skill is versatile enough to be immediately applicable to new shapes and target configurations.
\textsc{NutOnCone} requires to drop a nut onto a cone such that the cone is almost entirely hidden.
Conversely, \textsc{BallInCup} requires a ball to be dropped into a cup such that only a fraction of its surface remains visible.
Besides the handling of unseen object geometries, both tasks also pose novel challenges in terms of task inference because they feature much heavier occlusions than the original \textsc{Goal2Cube2} dataset which the model was trained on.
Our full model, $\mathcal{F}$, outperforms the ablations significantly in both new scenarios.
The encouraging results in both generalisation experiments shine additional light on GEECO's robustness and versatility and suggest that those capabilities can be achieved without resorting to expensive domain randomisation schemes during model training.

\section{Conclusions}\label{sec:conc}

We introduce GEECO, a novel architecture for goal-conditioned end-to-end visuomotor control utilising dynamic images.
GEECO can be immediately conditioned on a new task with the input of a single target image.
We demonstrate GEECO's efficacy in complex pushing and pick-and-place tasks involving multiple objects.
It also generalises well to challenging, unseen scenarios maintaining strong task performance even when confronted with heavy clutter, visual distortions or novel object geometries.
Additionally, its built-in invariances can help to reduce the dependency on sophisticated randomisation schemes during the training of visuomotor controllers.
Our results suggest that GEECO can serve as a robust component in robotic manipulation setups providing data-efficient and versatile skill primitives for manipulation of rigid objects.


\section*{ACKNOWLEDGMENTS}
Oliver Groth is funded by the European Research Council under grant ERC 638009-IDIU.
Chia-Man Hung is funded by the Clarendon Fund and receives a Keble College Sloane Robinson Scholarship at the University of Oxford.
This work is also supported by an EPSRC Programme Grant (EP/M019918/1).
The authors acknowledge the use of Hartree Centre resources in this work. TheSTFC Hartree Centre is a research collaboratory in association with IBM providing High Performance Computing platforms funded by the UK's investment in e-Infrastructure.
The authors also acknowledge the use of the University of Oxford Advanced Research Computing (ARC) facility in carrying out this work (\texttt{http://dx.doi.org/10.5281/zenodo.22558}).
Special thanks goes to Frederik Ebert for his helpful advise on adjusting Visual Foresight to our scenarios and to \cb{S}tefan S\u{a}ftescu for lending a hand in managing experiments on the compute clusters.
Lastly, we would also like to thank our dear colleagues Sudhanshu Kasewa, S\'ebastien Ehrhardt and Olivia Wiles for proofreading and their helpful suggestions and discussions on this draft.



\bibliographystyle{IEEEtran}
\bibliography{IEEEabrv,geeco}

\appendix
\subsection{GEECO Hyper-parameters}
\label{sec:geeco-hparams}
In this section, we present additional details regarding the architecture and training hyper-parameters of GEECO and all its ablations.

\paragraph{Observation Buffer.}
The observation buffer consists of pairs $(I_j,\mathbf{x}_j),~j~\in~[t-K+1,\dots,t]$ of images $I_j$ and proprioceptive features $\mathbf{x}_j$ representing the $K$ most recent observations of the model up to the current time step $t$.
The images are RGB with a resolution of $256 \times 256$ and the proprioceptive feature is a vector of length seven containing the angles of the robot's seven joints at the respective time step.
We have experimented with frame buffer sizes $K \in \{2, 4, 6, 8\}$.
Buffer sizes smaller than four result in too coarse approximations of dynamics (because velocities have to be inferred from just two time steps) and consequently in lower controller performance.
However, controller performance also does not seem to improve with buffer sizes greater than four.
We assume that in our scenarios, four frames are sufficient to capture the robot's motions accurately enough, which is in line with similar experiments in prior work~\cite{james2017corl}.
Therefore, we keep the buffer hyper-parameter $K = 4$ fixed in all our experiments.
At the start of the execution of the controller, we pad the observation buffer to the left with copies of the oldest frame, if there are less than $K$ pairs in the buffer assuming that the robot is always starting from complete rest.

\paragraph{Convolutional Encoder.}
All convolutional encoders used in the GEECO architecture have the same structure, which is outlined in \cref{table:suppmat-conv-encoder}.
However, the parameters between the convolutional encoders are not shared. The rationale behind this decision is that the different stacks of convolutions are processing semantically different inputs:
$\psi_{OBS}$ processes raw RGB observations, $\psi_{DYN}$ processes dynamic images representing the motion captured in the observation buffer and $\psi_{DIFF}$ processes the dynamic image difference between the current observation and the target image.

\begin{table}[h!]
\vskip 0.15in
\begin{center}
\begin{small}
\begin{sc}
\begin{tabular}{l|cccc}
\toprule
Layer & Filters & Kernel & Stride & Activation \\
\hline
Conv1 & 32 & 3 & 1 & ReLU \\
Conv2 & 48 & 3 & 2 & ReLU \\
Conv3 & 64 & 3 & 2 & ReLU \\
Conv4 & 128 & 3 & 2 & ReLU \\
Conv5 & 192 & 3 & 2 & ReLU \\
Conv6 & 256 & 3 & 2 & ReLU \\
Conv7 & 256 & 3 & 2 & ReLU \\
Conv8 & 256 & 3 & 2 & ReLU \\
\bottomrule
\end{tabular}
\end{sc}
\end{small}
\end{center}
\vskip 0.1in
\caption{The convolutional encoders used in GEECO all share the same structure of eight consecutive layers of 2D convolutions. They take as inputs RGB images with a resolution of $256 \times 256$ and return spatial feature maps with a shape of $2 \times 2 \times 256$.}
\label{table:suppmat-conv-encoder}
\end{table}

\paragraph{LSTM Decoder.}
The spatial feature maps $\psi_{OBS}(I_t)$, $\psi_{DYN}(\hat{\rho}(I_{t-K+1},\dots,I_t))$, $\psi_{DIFF}(\hat{\rho}(I_t, I_T))$ obtained from the convolutional encoders are concatenated to the proprioceptive feature $\mathbf{x}_t$ containing the current joint angles for the robot's 7 DoF.
This concatenated tensor forms the state representation $\mathbf{s}_t$, which, in the full model GEECO-$\mathcal{F}$, has a shape of $2 \times 2 \times (256 + 256 + 7 + 256)$.
The state is subsequently fed into an LSTM (cf.~\cref{fig:method-model}).
The LSTM has a hidden state $\mathbf{h}$ of size 128 and produces an output vector $\mathbf{o}_t$ of the same dimension at each time step.
As shown in prior work~\cite{james2017corl}, maintaining an internal state in the network is crucial for performing multi-stage tasks such as pick-and-place.

At the beginning of each task, i.e. when the target image $I_T$ is set and before the first action is executed, the LSTM state is initialised with a zero vector.
The output $\mathbf{o}_t$ at each timestep is passed through a fully connected layer $\phi(\cdot)$ with 128 neurons and a \textsc{ReLU} activation function.
This last-layer feature $\phi(\mathbf{o}_t)$ is finally passed through four parallel, fully-connected decoding heads without an activation function to obtain the command vectors and the auxiliary position estimates for the object and the end effector as described in~\cref{table:suppmat-lstm-decoder}.

\begin{table}[h!]
\vskip 0.15in
\begin{center}
\begin{small}
\begin{tabular}{l|cl}
\toprule
\textsc{Head} & \textsc{Units} & \textsc{Output} \\
\hline
$\mathbf{\hat{u}}_{\Delta EE}$ & 3 & change in EE position $(\Delta x, \Delta y, \Delta z)$ \\
$\mathbf{\hat{u}}_{GRP}$ & 3 & logits for \{open, noop, close\} \\
$\mathbf{\hat{q}}_{EE}$ & 3 & absolute EE position $(x, y, z)$ \\
$\mathbf{\hat{q}}_{POS}$ & 3 & absolute OBJ position $(x, y, z)$ \\
\bottomrule
\end{tabular}
\end{small}
\end{center}
\vskip 0.1in
\caption{The output heads of the LSTM decoder regressing to the commands and auxiliary position estimates.}
\label{table:suppmat-lstm-decoder}
\end{table}

\paragraph{Training Details.}
We train all versions of GEECO with a batch size of 32 for 300k gradient steps using the Adam optimiser~\cite{kingma2015iclr} with a start learning rate of 1e-4.
One training run takes approximately 48 hours to complete using a single NVIDIA GTX 1080 Ti with 11 GB of memory.

\paragraph{Execution Time.}
Running one simulated trial with an episode length of eight seconds takes about ten seconds for any version of GEECO using a single NVIDIA GTX 1080 Ti.
This timing includes the computational overhead for running and rendering the physics simulation resulting in a lower-bound estimate of GEECO's control frequency at 20 Hz.
This indicates that our model is nearly real-time capable of continuous control without major modifications.

\subsection{GEECO Ablation Details}
\label{sec:geeco-ablations}

\begin{figure}[h!]
    \centering
    \includegraphics[width=\columnwidth]{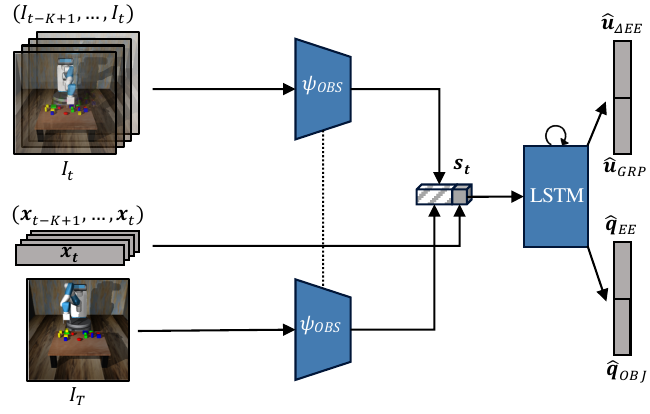}
    \caption{
    Model architecture of GEECO-$\mathcal{R}$.
    The same encoder $\psi_{OBS}$ is used for RGB observations $I_t$ and the target frame $I_T$.
    For each observed image $I_t$, the residual feature w.r.t. to the target image $I_T$ is computed as $\psi_{OBS}(I_T) - \psi_{OBS}(I_t)$, indicated by the striped box in $\mathbf{s}_t$.
    }
    \label{fig:suppmat-geeco_r}
\end{figure}

\paragraph{GEECO-$\mathcal{R}$}
Our first ablation, which is presented in \cref{fig:suppmat-geeco_r}, uses a na\"ive \emph{residual} target encoding to represent the distance to a given target observation in feature space.
The residual feature is the difference $\psi_{OBS}(I_T)-\psi_{OBS}(I_j),~j~\in~[t-K+1,\dots,t]$ and should tend towards zero as the observation $I_j$ approaches the target image $I_T$.
Since the same encoder $\psi_{OBS}$ is used for observation and target image, this architecture should encourage the formation of a feature space which captures the difference between an observation and the target image in a semantically meaningful way.
Since the observation buffer is not compressed into a dynamic image via $\hat{\rho}(\cdot)$, it is processed slightly differently in order to retain information about the motion dynamics.
For each pair $(I_j,\mathbf{x}_j),~j~\in~[t-K+1,\dots,t]$ containing an observed image and a proprioceptive feature at time step $j$, the corresponding state representation $\mathbf{s}_j$ is computed and fed into the LSTM which, in turn, updates its state.
However, only after all $K$ pairs of the observation buffer have been fed, the command outputs are decoded from the LSTM's last output vector.
This delegates the task of inferring motion dynamics to the LSTM as it processes the observation buffer.

\begin{figure}[h!]
    \centering
    \includegraphics[width=\columnwidth]{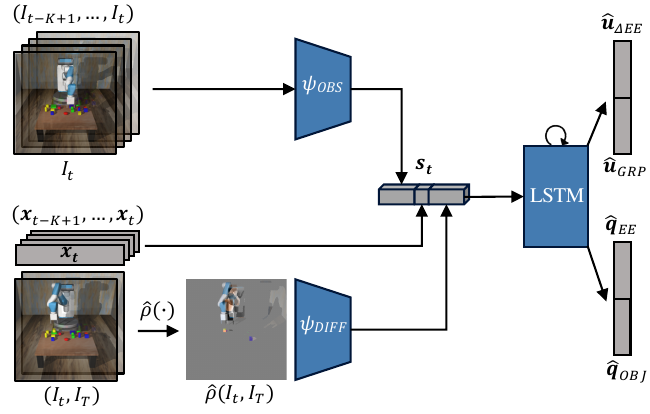}
    \caption{
    Model architecture of GEECO-$\mathcal{D}$.
    For each image $I_t$ in the observation buffer, the \emph{dynamic difference} to the target image $I_T$ is computed using $\hat{\rho}(\cdot)$.
    The difference image $\hat{\rho}(I_t,I_T)$ is encoded with $\psi_{DIFF}$ before being concatenated to $\mathbf{s}_t$.
    }
    \label{fig:suppmat-geeco_d}
\end{figure}

\paragraph{GEECO-$\mathcal{D}$}
Our second ablation, which is presented in \cref{fig:suppmat-geeco_d}, uses the dynamic image operator $\hat{\rho(\cdot)}$ to compute the difference between each observed frame $I_j,~j~\in~[t-K+1,\dots,t]$ and the target image $I_T$ as opposed to GEECO-$\mathcal{R}$ which represents the difference only in feature space.
Since the \emph{dynamic difference} $\hat{\rho}(I_t,I_T)$ is semantically different from a normal RGB observation, it is processed with a dedicated convolutional encoder $\psi_{DIFF}$ and the resulting feature is concantenated to the state representation $\mathbf{s}_t$.
In order to also capture motion dynamics, the observation buffer is processed sequentially like in GEECO-$\mathcal{R}$ before a control command is issued.

\subsection{E2EVMC Baseline}
\label{sec:e2evmc-baseline}
We compare GEECO to E2EVMC~\cite{james2017corl}, an unconditioned visuomotor controller, which we have implemented according to the original paper.
We have re-created a similar environment like in the original paper featuring only a red cube and a blue target pad which we call \textsc{Goal1Cube1}.
This dataset also consists of 4,000 demonstrations per skill and is split into training, validation and test sets with a ratio of 2 : 1 : 1.
Training and testing on this scenario is done to ensure the correct functionality of the model architecture and verify that GEECO performs at least as well as an unconditioned controller in an unambiguous scenario.
Even though the task is always the same, i.e. the red cube always goes on top of the blue pad, we still provide GEECO with the target image in every trial.
The unconditioned baseline, E2EVMC, runs without a target image since it is trained to perform only one task.
We train E2EVMC exactly like GEECO (cf.~\cref{sec:geeco-hparams}: Training Details) and select the best model snapshots according to the task performance on the respective validation sets.
We present experimental results for 1,000 trials on the test set of \textsc{Goal1Cube1} in~\cref{table:suppmat-e2evmc}.

\begin{table*}[h!]
\vskip 0.15in
\begin{center}
\begin{small}
\begin{sc}
\begin{tabular}{l|cc|ccc}
\toprule
\multirow{3}{*}{model} & \multicolumn{2}{c|}{Pushing}  & \multicolumn{3}{c}{Pick-and-Place} \\
& reach & push & reach & pick & place \\
& {[}\%{]} & {[}\%{]} & {[}\%{]} & {[}\%{]} & {[}\%{]} \\
\hline
E2EVMC~\cite{james2017corl} & 95.20 $\pm$ 1.32 & 58.60 $\pm$ 3.05 & \textbf{96.10 $\pm$ 1.20} & 72.00 $\pm$ 2.78 & \textbf{67.60 $\pm$ 2.90} \\
\hline
$\mathcal{R}$ & \textbf{98.80 $\pm$ 0.67} & 43.70 $\pm$ 3.07 & \textbf{95.30 $\pm$ 1.31} & 73.00 $\pm$ 2.75 & 60.70 $\pm$ 3.03 \\
$\mathcal{D}$ & \textbf{99.50 $\pm$ 0.44} & \textbf{87.40 $\pm$ 2.06} & \textbf{95.80 $\pm$ 1.24} & 77.40 $\pm$ 2.59 & \textbf{64.90 $\pm$ 2.96} \\
$\mathcal{F}$ & \textbf{99.20 $\pm$ 0.55} & 72.90 $\pm$ 2.75 & \textbf{95.90 $\pm$ 1.23} & \textbf{83.30 $\pm$ 2.31} & 61.20 $\pm$ 3.02 \\
\bottomrule
\end{tabular}
\end{sc}
\end{small}
\end{center}
\vskip 0.1in
\caption{
Comparison of pushing and pick-and-place performance of all versions of GEECO with E2EVMC on \textsc{Goal1Cube1}.
}
\label{table:suppmat-e2evmc}
\end{table*}

We observe that GEECO-$\mathcal{D}$ and -$\mathcal{F}$ perform commensurately with E2EVMC for both pushing and pick-and-place tasks.
Both E2EVMC and GEECO reach the red cube nearly perfectly with at least $95\%$ success rate.
GEECO-$\mathcal{D}$ performs best on this dataset even outperforming E2EVMC by almost $30\%$ mean success rate for pushing tasks.
Again, GEECO-$\mathcal{F}$ sometimes exhibits its failure mode at the pivot point between moving and dropping phase presumably due to ambiguous or uninformative signals from the motion representation around this phase.

\subsection{Visual Foresight Baseline}
\label{sec:vfs-hparams}
In this section, we explain all hyper-parameters which have been used during training and evaluation of the Visual Foresight baseline~\cite{ebert2018arxiv}.


\paragraph{Video Predictor.}
We use the official implementation\footnote{\url{https://github.com/alexlee-gk/video_prediction}} of Stochastic Adversarial Video Prediction (SAVP)~\cite{lee2018savp} as the video prediction backbone of Visual Foresight.
We have not been able to fit the model at a resolution of $256 \times 256$ on a single GPU with 11 GB of memory.
Hence, we adjusted the image resolution of the video predictor to $128 \times 128$ pixels.
We use SAVP's hyper-parameter set which is reported for the BAIR robot pushing dataset~\cite{ebert2017corl} since those scenarios resemble our training setup most closely.
We report the hyper-parameter setup in~\cref{table:suppmat-savp-hparams}.
\begin{table}[h!]
\vskip 0.15in
\begin{center}
\begin{small}
\begin{tabular}{l|cl}
\toprule
\textsc{Parameter} & \textsc{Value} & \textsc{Description} \\
\hline
\texttt{scale\_size} & $128$ & image resolution \\
\texttt{use\_state} & True & use action conditioning \\
\texttt{sequence\_length} & $13$ & prediction horizon \\
\texttt{frame\_skip} & $0$ & use entire video \\
\texttt{time\_shift} & $0$ & use original frame rate \\
\texttt{l1\_weight} & $1.0$ & use $L_1$ reconstruction loss \\
\texttt{kl\_weight} & $0.0$ & make model deterministic \\
\texttt{state\_weight} & 1e-4 & weight of conditioning loss \\
\bottomrule
\end{tabular}
\end{small}
\end{center}
\vskip 0.1in
\caption{Hyper-parameter setup of SAVP. Hyper-parameters not listed here are kept at their respective default values.}
\label{table:suppmat-savp-hparams}
\end{table}

\paragraph{Training Details.}
We train SAVP with a batch size of 11 for 300k gradient steps using the Adam optimiser~\cite{kingma2015iclr} with a start learning rate of 1e-4.
One training run takes approximately 72 hours to complete using a single NVIDIA GTX 1080 Ti with 11 GB of memory.

\paragraph{Action Sampling.}
We use CEM~\cite{rubinstein2004springer} as in the original VFS paper~\cite{ebert2018arxiv} to sample actions which bring the scene closer to a desired target image under the video prediction model.
We set the \emph{planning horizon} of VFS to the prediction length of SAVP, $P = 13$.
The action space is identical to the one used in GEECO and consists of a continuous vector representing the position change in the end effector $\mathbf{u}_{\Delta EE} \in \mathbb{R}^3$ and a discrete command for the gripper $\mathbf{u}_{GRP} \in \{-1,0,1\}$.
Once a target image has been set, we sample action sequences of length $P$ according to the following \cref{eq:vfs-u_dee,eq:vfs-u_grp}:
\begin{align}
&\mathbf{u}_{\Delta EE}^{1 \colon P} \sim \mathcal{N}(\mathbf{\mu},\Sigma)
\label{eq:vfs-u_dee}\\
&\mathbf{u}_{GRP}^{1 \colon P} \sim \mathcal{U}\{-1,0,1\}
\label{eq:vfs-u_grp}
\end{align}
where $\mathcal{N}(\mathbf{\mu},\Sigma)$ is a multi-variate Gaussian distribution and $\mathcal{U}\{-1,0,1\}$ is a uniform distribution over the gripper states.
For each planning step, we run CEM for four iterations drawing 200 samples at each step and re-fit the distributions to the ten best action sequences according to the video predictor, i.e. the action sequences which transform the scene closest to the next goal image.
Finally, we execute the best action sequence yielded from the last CEM iteration and re-plan after $P$ steps.

\begin{figure}[h!]
  \includegraphics[width=\columnwidth]{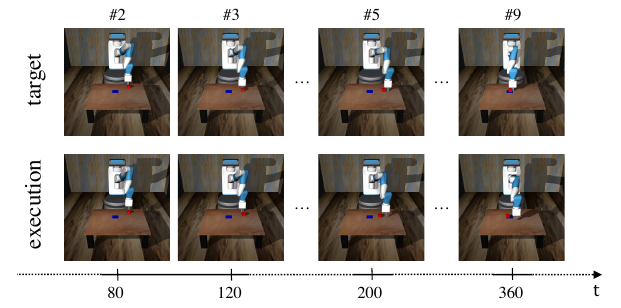}
  \caption{
  An execution of VFS with the `ground truth bottleneck' scheme.
  The top row depicts intermediate target images from an expert demonstration.
  The bottom row shows the corresponding state of execution via VFS at time step $t$.
  }
  \label{fig:suppmat-vfs-goals}
\end{figure}

\paragraph{Goal Distance.}
We use $L_2$ distance in image space to determine the distance between an image forecast by the video predictor and a target image (cf.~\cite{ebert2018arxiv}).
Since this goal distance is dominated by large image regions (e.g. the robot arm), it is ill suited to capture position differences of the comparatively small objects on the table or provide a good signal when a trajectory is required which is not a straight line.
Therefore, we resort to a `ground truth bottleneck' scheme~\cite{nair2019arxiv} for a fairer comparison.
Instead of providing just a single target image from the end of an expert demonstration, we give the model ten \emph{intermediate target frames} taken every ten steps during the expert demonstration.
This breaks down the long-horizon planning problem into multiple short-horizon ones with approximately straight-line trajectories between any two intermediate targets.
This gives an upper-bound estimate of VFS's performance, if it had access to a perfect keyframe predictor splitting the long-horizon problem.
An example execution of VFS being guided along intermediate target frames is presented in \cref{fig:suppmat-vfs-goals}.

\begin{table*}[t]
\vskip 0.15in
\begin{center}
\begin{small}
\begin{tabular}{l|cl}
\toprule
\textsc{Parameter} & \textsc{Value} & \textsc{Description} \\
\hline
\texttt{iterations} & $300000$ & number of gradient updates \\
\texttt{batch\_size} & $64$ & batch size \\
\texttt{lr} & 5e-4 & start learning rate of Adam optimiser \\
\texttt{img\_shape} & (125, 125, 3) & image resolution \\
\texttt{support} & $1$ & k-shot support of new task \\
\texttt{query} & $5$ & query examples per task during training \\
\texttt{embedding} & $20$ & size of task embedding vector \\
\texttt{activation} & \textsc{elu} & layer activation function \\
\texttt{filters} & 16,16,16,16 & number of filters in each conv-layer of the embedding network \\
\texttt{kernels} & 5,5,5,5 & filter size in each conv-layer of the embedding network \\
\texttt{strides} & 2,2,2,2 & stride size in each conv-layer of the embedding network \\
\texttt{fc\_layers} & 200,200,200 & neurons of the fc-layers of the control network \\
\texttt{lambda\_embedding} & $1.0$ & weight of embedding loss \\
\texttt{lambda\_support} & $0.1$ & weight of support loss \\
\texttt{lambda\_query} & $0.1$ & weight of query loss \\
\texttt{margin} & $0.1$ & margin of the hinge rank loss \\
\texttt{norm} & \textsc{layer} & using Layer-Norm throughout the network \\
\bottomrule
\end{tabular}
\end{small}
\end{center}
\vskip 0.1in
\caption{Hyper-parameter setup of \textsc{TecNet}. Hyper-parameters not listed here are kept at their respective default values.}
\label{table:suppmat-tecnet-hparams}
\end{table*}

\paragraph{Execution Time.}
To account for VFS's sampling-based nature and the guided control process using intermediate target images, we give VFS some additional time to execute a task during test time.
We set the total test episode length to 400 time steps as opposed to 200 used during the evaluation of GEECO.
VFS is given 40 time steps to `complete' each sub-goal presented via the ten intermediate target images.
However, the intermediate target image is updated to the next sub-goal strictly every 40 time steps, irrespective of how `close' the controller has come to achieving the previous sub-goal.
Running one simulated trial with an episode length of 16 seconds takes about ten minutes using a single NVIDIA GTX 1080 Ti.
This timing includes the computational overhead for running and rendering the physics simulation.
While this results in an effective control frequency of 0.7 Hz, a like-for-like comparison between VFS and GEECO can not be made in that regard because we have not tuned VFS for runtime efficiency in our scenarios.
Potential speedups can be gained from lowering the image resolution and frame rate of the video predictor, predicting shorter time horizons and pipelining the re-planning procedure in a separate thread.
However, the fundamental computational bottlenecks of visual MPC can not be overcome with hyper-paramter tuning:
Action-conditioned video prediction remains an expensive operation for dynamics forecasting although pixel-level prediction accuracy is presumably not needed to control a robot.
Additionally, the action sampling process is a separate part of the model which requires tuning and trades off accuracy versus execution time.
In contrast to that, GEECO provides a compelling alternative by reducing the action computation to a single forward pass through the controller network.

\subsection{TecNet Baseline}
\label{sec:tecnet-hparams}

We use the official implementation of \textsc{TecNet} for all our experiments\footnote{\url{https://github.com/stepjam/TecNets}}.
In~\cref{table:suppmat-tecnet-hparams} we provide a comprehensive list of all hyper-parameters used in our experiments with \textsc{TecNet}.

\paragraph{Training Details.}
We train \textsc{TecNet} in the one-shot imitation setup and provide one `demonstration' before the start of the controller execution consisting of only the first observation and the target image.
During training and evaluation, we resize all images fed into \textsc{TecNet} to $125 \times 125$ pixels as per the original paper.
We train \textsc{TecNet} with a batch size of 64 for 300k gradient update steps using the Adam optimiser~\cite{kingma2015iclr} with a start learning rate of 5e-4.
One training run takes about 72 hours to complete using a single NVIDIA GTX 1080 Ti with 11 GB of memory.

\paragraph{Execution Time.}
Running one simulated trial with \textsc{TecNet} with an episode length of eight seconds takes about eight seconds using a single NVIDIA GTX 1080 Ti.
This timing includes the computational overhead for running and rendering the physics simulation resulting in a lower-bound estimate of \textsc{TecNet}'s control frequency at 25 Hz.
This makes \textsc{TecNet} also a viable option for real-time visuomotor control without any system modifications.

\end{document}